\documentclass[10pt,journal,compsoc]{IEEEtran}

\ifCLASSOPTIONcompsoc
\usepackage[nocompress]{cite}
\else
\usepackage{cite}
\fi
\ifCLASSINFOpdf
\usepackage[pdftex]{graphicx}
\else
\fi

\usepackage{amsmath}
\usepackage{algorithm}
\usepackage{algorithmic}
\usepackage{array}
\ifCLASSOPTIONcompsoc
\usepackage[caption=false,font=footnotesize,labelfont=sf,textfont=sf]{subfig}
\else
\usepackage[caption=false,font=footnotesize]{subfig}
\fi

\usepackage{stfloats}
\usepackage{url}

\usepackage{float}
\usepackage{amssymb}
\usepackage{slashbox}
\usepackage{epsfig}
\usepackage{graphicx}
\usepackage{babel}
\usepackage{threeparttable}
\usepackage{booktabs}
\usepackage{makecell}
\usepackage{diagbox}
\usepackage{multirow}
\usepackage{color}
\usepackage[colorlinks,linkcolor=blue,anchorcolor=blue,citecolor=blue]{hyperref}

\newcommand{\tabincell}[2]{\begin{tabular}{@{}#1@{}}#2\end{tabular}}

\begin{document}
\title{FingerGAN: A Constrained Fingerprint Generation Scheme for Latent Fingerprint Enhancement
}

\author{
        Yanming~Zhu,
        Xuefei~Yin,
        and~Jiankun~Hu$^{*}$,~\IEEEmembership{Senior Member,~IEEE}
\IEEEcompsocitemizethanks{\IEEEcompsocthanksitem Yanming Zhu is with the School of Computer Science and Engineering, University of New South Wales, Sydney, NSW 2052, Australia (e-mail: yanming.zhu@unsw.edu.au).
\IEEEcompsocthanksitem Xuefei Yin and Jiankun Hu are with School of Engineering and Information Technology, University of New South Wales, Canberra, ACT 2600, Australia (e-mail:  xuefei.yin@unsw.edu.au; $^*$Corresponding author: J.Hu@adfa.edu.au).}
}

\IEEEtitleabstractindextext{
\begin{abstract}
Latent fingerprint enhancement is an essential pre-processing step for latent fingerprint identification. Most latent fingerprint enhancement methods try to restore corrupted gray ridges/valleys. In this paper, we propose a new method that formulates the latent fingerprint enhancement as a constrained fingerprint generation problem within a generative adversarial network (GAN) framework. We name the proposed network as FingerGAN. It can enforce its generated fingerprint (i.e, enhanced latent fingerprint) indistinguishable from the corresponding ground-truth instance in terms of the fingerprint skeleton map weighted by minutia locations and the orientation field regularized by the FOMFE model. Because minutia is the primary feature for fingerprint recognition and minutia can be retrieved directly from the fingerprint skeleton map, we offer a holistic framework which can perform latent fingerprint enhancement in the context of directly optimizing minutia information. This will help improve latent fingerprint identification performance significantly. Experimental results on two public latent fingerprint databases demonstrate that our method outperforms the state of the arts significantly. The codes will be available for non-commercial purposes from \url{https://github.com/HubYZ/LatentEnhancement}.
\end{abstract}

}

\maketitle
\IEEEdisplaynontitleabstractindextext
\IEEEpeerreviewmaketitle

\IEEEraisesectionheading{\section{Introduction}}
\IEEEPARstart{F}{ingerprints} have been widely used for human verification and identification in many civil or criminal applications \cite{1HawthorneFingerprints,2MaltoniHandbook}.
Different from plain and rolled fingerprints that are acquired professionally, latent fingerprints refer to the finger skin impressions unintentionally left at a crime scene and are generally used as an important evidence to identify criminals by law enforcement and forensic agencies. Compared with plain and rolled fingerprints, latent fingerprints are usually smudgy and blurred, with incomplete regions, unclear ridge structures, and complex background noise. Due to these factors, the identification accuracy of latent fingerprints, which heavily relies on the fingerprint quality, is much lower than that of plain and rolled fingerprints. Therefore, latent fingerprint enhancement, which aims to improve latent fingerprint quality, becomes one of the most necessary and important pre-processing steps for latent fingerprint identification. It refers to the task of removing background noise, improving ridge-valley contrast, enhancing ridge and valley structures, and restoring corrupted structures. 

In the past decades, many efforts have been made towards the latent fingerprint enhancement \cite{wallace2004detection,sankaran2014latent,malwade2015survey,schuch2018survey,abebe2020latent}. At the early stage, classical image processing techniques such as contextual filtering and directional filtering were introduced to enhance fingerprints. For example, Cappelli et al. \cite{3cappelli2009semi} proposed tuning a Gabor filter to the local orientations and frequencies of fingerprints to suppress noise and improve the clarity of ridge structure. Chikkerur et al. \cite{4chikkerur2007fingerprint} proposed performing a contextual filtering in the Fourier domain to enhance fingerprints. However, these methods are mainly effective for bad-quality plain or rolled fingerprints, and tend to fail in latent fingerprint enhancement due to: 1) the corrupted ridge structures caused by structural noise in latent fingerprints; and 2) the unreliable orientation and frequency estimation caused by the low clarity of ridge structures of latent fingerprints. Therefore, varieties of smoothing and global modeling techniques were proposed to address the above problems and devoted to reliable orientation estimation to improve the latent fingerprint enhancement \cite{6yoon2010latent,7yoon2011latent,8feng2013orientation,9yang2014localized}. For example, Yoon et al. \cite{7yoon2011latent} proposed using a polynomial model together with Gabor filters to estimate the fingerprint orientations to improve the latent enhancement. Feng et al. \cite{8feng2013orientation} proposed using an orientation patch dictionary to estimate orientations and applying Gabor filtering to the orientations to achieve latent fingerprint enhancement. Yang et al. \cite{9yang2014localized} proposed to improve the above method by replacing its orientation dictionary with a set of localized orientation dictionaries. Because these localized dictionaries are different from each other according to their locations, more orientation patch candidates can be selected to estimate more reliable orientations and thus to improve the enhancement. However, due to the fixed ridge frequency in the tuning of Gabor filters, the performance of these methods is limited in reality where the ridge frequency of fingerprints is not constant.

Later, to further improve the enhancement of latent fingerprints, total variation (TV) image models, which minimize the total variation of an image and decompose the image into two components of texture and cartoon, were adopted to take advantage of ridge structures \cite{10zhang2012latent,11zhang2013adaptive,12cao2014segmentation,13liu2015latent}. For example, Zhang et al. firstly proposed an adaptive TV model \cite{10zhang2012latent} to remove the structural noise of latent fingerprints and then proposed an adaptive directional TV model \cite{11zhang2013adaptive} for latent fingerprint enhancement. These methods can restrain the structural noise in the decomposed texture components of latent fingerprints by integrating local orientations and scales of fingerprints. However, estimating local parameters of these models for poor-quality latent fingerprints is difficult and thus the extracted ridge structures by these models are usually weak. Therefore, in later researches, the TV decomposition is generally used as a pre-processing for latent fingerprint enhancement \cite{12cao2014segmentation,13liu2015latent}. 

After that, with the success of deep learning, deep neural networks were proposed for fingerprint tasks including latent fingerprint enhancement \cite{tang2017fingernet,nguyen2018robust}. For example, Cao et al. \cite{18cao2015latent} proposed a method which uses a convolutional neural network (CNN) to estimate the orientation field and then uses Gabor filter banks to obtain the final enhanced fingerprint. Svoboda et al. \cite{19svoboda2017generative} proposed using a convolutional autoencoder network to reconstruct latent fingerprints. Similarly, Li et al. \cite{20li2018deep} proposed a deep convolutional network consisting of one convolution and two deconvolution parts for latent fingerprint enhancement. Horapong et al. \cite{horapong2020progressive} proposed using a sparse autoencoder to boost the ridge/valley spectrum for latent fingerprint enhancement. Liu et al. \cite{liu2020automatic} proposed using a deep nested UNet for latent fingerprint enhancement. These methods take advantages of the strong representation ability of deep neural networks and achieve remarkable results, but the corrupted ridge/valley structures of latent fingerprints are not well restored in most cases. 

Recently, generative adversarial networks (GANs) are adopted to latent fingerprint enhancement to enhance the restoration of ridge/valley structures. For example, Dabouei et al. \cite{21dabouei2018id} proposed a conditional GAN for partial latent fingerprint enhancement, which achieves an enhancement of rejecting seriously corrupted fingerprint regions while improving ridge structure clarity in comparatively good-quality regions. Joshi et al. \cite{joshi2019latent} proposed a GAN-based algorithm to amplify the ridge/valley structure of latent fingerprint for the enhancement. Huang et al. \cite{huang2020latent} proposed using a progressive PatchGAN to achieve latent fingerprint enhancement. The enhancement ability of these methods mainly comes from the powerful feature representation and reconstruction ability of GANs.

In this paper, we propose a new method that formulates the latent fingerprint enhancement as a constrained fingerprint generation problem within a GAN framework. The proposed network is named as FingerGAN. It can enforce its generated fingerprint (i.e, enhanced latent fingerprint) indistinguishable from the corresponding ground-truth instance in terms of the fingerprint skeleton map weighted by minutia locations and the orientation field regularized by the FOMFE model. Because minutia is the primary feature for recognition and minutia can be retrieved directly from the fingerprint skeleton map, we offer a holistic framework which can perform latent fingerprint enhancement in the context of directly optimizing minutia information. This will help improve latent fingerprint identification performance significantly. Experimental results on two public latent fingerprint databases demonstrate that our method significantly outperforms the state of the arts.

The main contributions of this paper are summarized as follows.
\begin{itemize}
	\item[1)] Unlike most latent fingerprint enhancement methods that try to restore corrupted gray ridges/valleys, we propose a new method which formulates the latent fingerprint enhancement as a constrained fingerprint generation problem within a GAN framework.
	
	\item[2)]  We propose a FingerGAN which can generate enhanced latent fingerprints conditioned on a fingerprint-to-fingerprint translation and can enforce its generated enhanced latent fingerprints indistinguishable from the ground-truth instances in terms of fingerprint skeleton map and orientation field.
	
	\item [3)] The fingerprint skeleton map is proposed as a ground-truth because minutia is the primary feature for recognition and minutia can be retrieved directly from the fingerprint skeleton map. Also, a Gaussian-based minutia weight map is proposed to apply to the reconstruction loss, which can accommodate a moderate loss of the accuracy of minutia locations.
	
	\item[4)] The orientation field is proposed as a ground-truth in a way of bringing in correspondence between the generated enhanced latent fingerprint and the ground-truth orientation field. Also, the FOMFE model is adopted to regularize the orientation field so that the effects of spurious pixels and noise can be rectified. 
	
	\item [5)] A synthetic latent fingerprint generation method is proposed, which can address the issue of lacking high-volume latent fingerprints and their true mates required for deep learning.
	
\end{itemize}

The rest of this paper is organized as follows. Section \ref{sec:relatedworks} provides background information on related techniques. Section \ref{sec:proposedmethod} describes the proposed method in detail. Section \ref{sec:experimentalresults} presents and discusses the experimental results. Finally, the paper is concluded in Section \ref{sec:conclusion}.

\section{Background}
\label{sec:relatedworks}
Since the proposed method involves GAN, autoencoder (AE), and FOMFE fingerprint orientation model, the related background knowledge is provided as follows.

\subsection{Generative Adversarial Network}
\label{subsec:ganIntro}
GAN is one of the most popular groups of generative networks, which learns to map an embedding space to a data distribution of interest, and has achieved great success in various image generation and processing tasks \cite{21dabouei2018id,23reed2016generative,24isola2017image}. The underlying strategy of a GAN is emulating a competition, with a generative network, called generator $G$, which takes a random vector $z$ sampled from a noise distribution $\mathcal{Z}$ as input and tries to generate samples as `real' as possible, and a discriminative network, called discriminator $D$, which performs binary classification to distinguish samples generated by $G$ from the real samples and acts as an adversary. The goal of $G$ is to maximize the misclassification error of $D$ while the goal of $D$ is to beat $G$ by learning to identify the generated samples. Through such a zero-sum game, the GANs have the ability to learn any kind of data distribution in an unsupervised manner. The networks of $G$ and $D$ are trained iteratively with two steps: 1) fixing the parameters of $G$ and optimizing $D$; and 2) fixing the parameters of $D$ and optimizing $G$ by using a loss function formulated as \cite{RN1275}:
\begin{equation}
\label{eq:gan}
\begin{aligned}
\min_{G} \max_{D} L(G,D) & = \mathbb{E}_{x \in \mathcal{X}}[log(D(x))]  \\
& + \mathbb{E}_{z \in \mathcal{Z}}[log(1-D(G(z)))],
\end{aligned}
\end{equation}
where $x$ is the real sample from the data distribution $\mathcal{X}$. $D(x)$ represents the binary classification score given input $x$.
During the training, half of samples are real and the rest $G(z)$ are samples generated by $G$ given $z$. 

Although the superiority of GANs in unsupervised representation learning, it can not be directly used to a latent fingerprint enhancement task due to its high probability of generating unrelated fingerprints. A GAN conditioned on the given information or constrained by prior knowledge can address this issue \cite{2401yeh2017semantic,isola2017image}, which inspires us to propose a FingerGAN. It is elaborately designed by customizing a GAN to fit the latent fingerprint enhancement task. 

\begin{figure*}[b]
	\centering
	\includegraphics[width=0.8\linewidth]{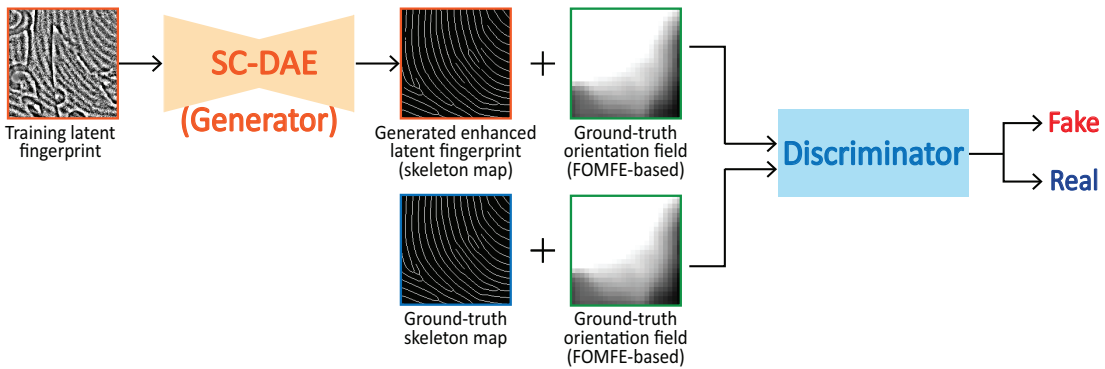}
	\caption{Framework of the proposed FingerGAN. Texture components from the TV decomposition \cite{RN761} are used as the latent fingerprints input to the SC-DAE because they are generally used as the representation of the latent fingerprints to be enhanced in current researches \cite{10zhang2012latent,11zhang2013adaptive,12cao2014segmentation,13liu2015latent}.}
	\label{fig:framework}
\end{figure*}

\subsection{Autoencoder (AE) and Its Variations}
\label{autencoderIntro}
AE is a popular tool to learn efficient data representation in an unsupervised manner \cite{25bengio2009learning,27hong2015multimodal,28yu2018multitask}. It is a type of artificial neural network and usually consists of an encoder that maps the input to a compact feature representation and a decoder that tries to generate a reconstruction from the compact representation as close as possible to the original input. It is trained by minimizing a reconstruction loss expressed as:
\begin{equation}
L(y,y') = ||y-y'||^2, 
\end{equation}
where $y$ is the input and $y'$ is its reconstruction from the AE. However, such an AE tends to compress the input to obtain a feature representation without the ability of learning semantically meaningful information. Therefore, denoising AE (DAE) is proposed to address this problem by corrupting the input and forcing the network to recover the original uncorrupted one. DAE has been widely used to extract more robust feature representations in various applications \cite{29vincent2008extracting,30vincent2010stacked,31lu2013speech}. To train a DAE, it is necessary to perform a preliminary stochastic mapping $y \rightarrow \tilde{y}$ to corrupt the input $y$ and use $\tilde y$ as the input to the network, with the loss still being computed for the initial input $y$ as:
\begin{equation}
L(y, \tilde{y}') = || y- \tilde{y}' ||^2,
\end{equation}
where $\tilde{y}'$ is the reconstruction from the DAE. Generally, the corruption process for a DAE is localized and low-level, thus the DAE does not require much semantic information to recover. For a task that requires a much deeper semantic understanding of the input, Pathak et al. \cite{32pathak2016context} proposed to train an AE with a loss function defined to jointly minimize a reconstruction loss and an adversarial loss, which achieved remarkable results. 

Therefore, inspired by this, we propose embedding a skip-connected DAE (SC-DAE) in a GAN for the latent fingerprint enhancement task. It can leverage the advantage of the DAE by using rolled fingerprints and their corresponding synthesized latent fingerprints to mimic the corruption mapping, and the advantage of GAN by jointly training the SC-DAE with an adversarial loss as in \cite{32pathak2016context}. 

\subsection{FOMFE Model}
\label{subsec:fomfe}
FOMFE model describes the global topology of fingerprint ridges and is for modeling fingerprint orientations \cite{39wang2007fingerprint}. It has the ability of predicting orientations for partial fingerprints and works well for low-quality fingerprints. Therefore, in this paper, we introduce it as prior knowledge to guide the constrained fingerprint generation.

\section{Proposed Method}
\label{sec:proposedmethod}
\subsection{Problem Formulation}
\label{subsec:problemformulation}
We propose to formulate the latent fingerprint enhancement as a constrained fingerprint generation problem conditioned on a fingerprint-to-fingerprint translation. For this purpose, we propose a FingerGAN by embedding a skip-connected DAE (SC-DAE) in a GAN such that the SC-DAE acts as the generator of the GAN, as illustrated in Fig. \ref{fig:framework}. 

The SC-DAE is responsible for generating enhanced latent fingerprints given input latent fingerprints. Because minutia is the primary feature for recognition and minutia can be retrieved directly from the fingerprint skeleton map, we propose using minutia location weighted fingerprint skeleton map as a ground-truth to force the SC-DAE performing latent fingerprint enhancement in the context of directly optimizing minutia information. The discriminator is used to enforce the SC-DAE generating enhanced latent fingerprints indistinguishable from the ground-truth instances in terms of both the fingerprint skeleton map and the FOMFE-based orientation field. For this purpose, its input is a concatenation\footnote{A concatenation is a two-channel tensor where the fingerprint skeleton map is its first channel map and the FOMFE-based orientation field is its second channel map.} of the fingerprint skeleton map and the FOMFE-based orientation field. Specifically, the SC-DAE generated enhanced latent fingerprint and the ground-truth orientation field are concatenated to form a type of input. The ground-truth skeleton map and the ground-truth orientation field are concatenated to form another type of input. The discriminator tries to distinguish these two types of inputs to beat the SC-DAE. This design of concatenation brings in correspondence between the generated enhanced fingerprint and the ground-truth orientation field. Therefore, the generation of the SC-DAE is constrained by prior knowledge of FOMFE-based orientation field and can address the problem of generating unrelated fingerprints. Details of the proposed FingerGAN are provided in the following Section \ref{subsec:aa}.  

\subsection{Details of the Proposed FingerGAN}
\label{subsec:aa}

\begin{figure*}[htbp]
	\centering
	\includegraphics[width=\linewidth]{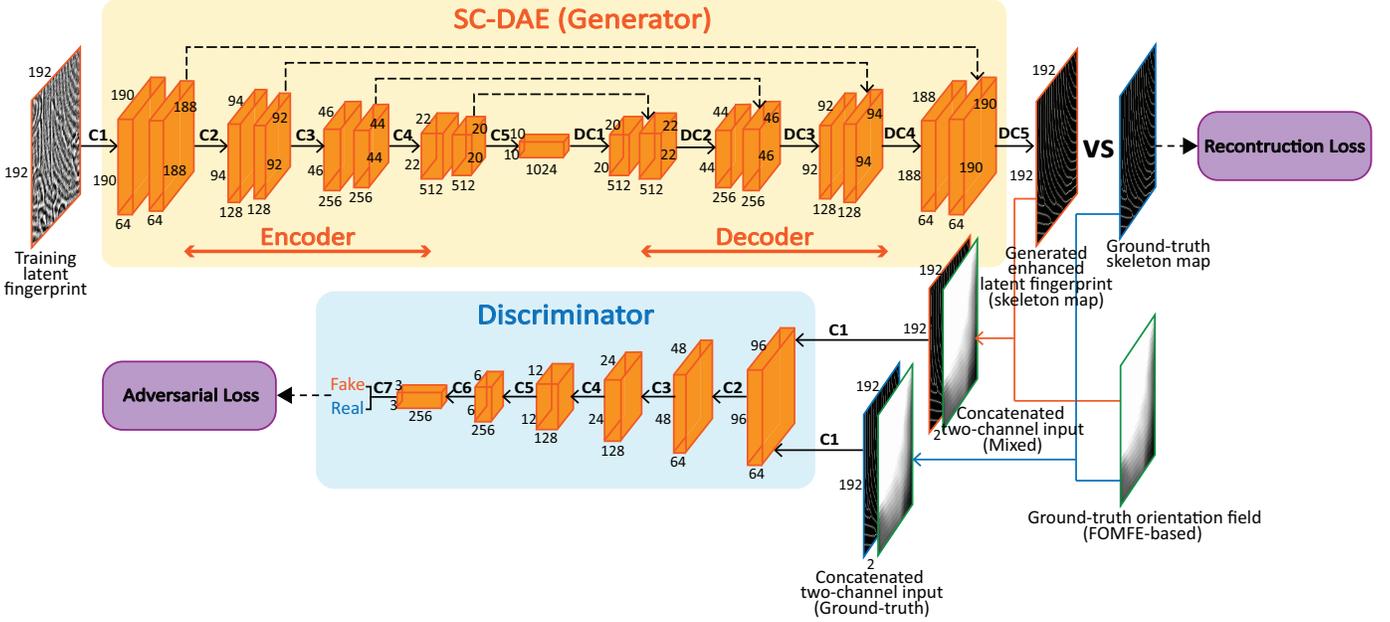}
	\vspace{-0.2in}
	\caption{Illustration of details of the proposed FingerGAN.}
	\label{fig:aa}
\end{figure*}

Fig. \ref{fig:aa} illustrates the details of the proposed FingerGAN.
\subsubsection{SC-DAE}
The SC-DAE consists of an encoder with five composite convolutional blocks (C1-C5) and a decoder with five deconvolutional blocks (DC1-DC5), where skip connection \cite{16mao2016image} is adopted for the first four composite deconvolutional blocks. This is proposed to keep the high frequency details of the inputs and increase the quality of the reconstruction from the decoder. Each of the first four composite convolutional blocks consists of two convolutional layers, and each convolutional layer is followed by a batch-normalization layer and a leaky rectified linear unit (ReLU) \cite{44DahlImproving}. The last composite convolutional block consists of one convolutional layer which is followed by a batch-normalization layer and a leaky ReLU layer. Each of the first four composite deconvolutional blocks consists of two up-convolutional layers, and each up-convolutional layer is followed by a batch-normalization layer and a leaky ReLU layer. The last composite deconvolutional block consists of an up-convolutional layer, a batch-normalization layer, and a sigmoid layer. According to the study in \cite{37simonyan2014very}, successive convolutions by a set of small kernels equal to one convolution by a lager kernel. It can effectively enhance a network's discriminative power and reduce the number of parameters required to be learned. In this paper, we use a set of small kernels and their details are reported in Table \ref{table:parameters}. Also, we double or halve the kernel numbers when the size of feature maps halving or doubling.

The input to the SC-DAE is a latent fingerprint to be enhanced, and the output of the SC-DAE is the generated corresponding enhanced latent fingerprint. In the training stage, the input latent fingerprints are synthesized using rolled fingerprints by our proposed method described in the following Section \ref{subsec:dataprepare}. The ground-truths used to optimize the generated enhanced latent fingerprints are the fingerprint skeleton maps of the rolled fingerprints. This way, by calculating a reconstruction loss between the generated enhanced latent fingerprint and the ground-truth, the SC-DAE can learn to de-noise the input latent fingerprint and reconstruct its fingerprint skeleton. 

\begin{table*}[htbp]
	\renewcommand{\arraystretch}{1.2}
	\caption{Details of the architecture of the FingerGAN}
	\vspace{-0.05in}
	\label{table:parameters}
	\centering
	\begin{tabular}{|@{\hspace{0.3em}}c@{\hspace{0.3em}}|c|c|@{\hspace{0.2em}}c@{\hspace{0.2em}}|@{\hspace{0.3em}}c@{\hspace{0.3em}}|c@{\hspace{0.01em}}|@{\hspace{0.3em}}c@{\hspace{0.3em}}|c|c|@{\hspace{0.2em}}c@{\hspace{0.2em}}|@{\hspace{0.3em}}c@{\hspace{0.3em}}|c@{\hspace{0.01em}}@{\hspace{0.05em}}c@{\hspace{0.05em}}|c|c|@{\hspace{0.2em}}c@{\hspace{0.2em}}|c|}
		\cline{1-11} \cline{13-17}
		\multicolumn{11}{|c|}{SC-DAE} &&\multicolumn{5}{|c|}{Discriminator} \\ 
        \cline{1-11} \cline{13-17}
		Block & Layer & \tabincell{c}{Kernel \\Size} & Stride &  \tabincell{c}{Kernel \\Number} && Block & Layer & \tabincell{c}{Kernel \\Size} & Stride &  \tabincell{c}{Kernel \\Number}&& \multicolumn{1}{|c|}{Block}  & Layer & \tabincell{c}{Kernel \\Size} & Stride &  \tabincell{c}{Kernel \\Number}\\
		\cline{1-5} \cline{7-11} \cline{13-17}
		\multirow{2}{*}{C1} & conv1 & 3$\times$3 & 1 &64 && \multirow{2}{*}{DC1} & up-conv1 & 2$\times$2 & 2 &512 && \multicolumn{1}{|c|}{C1} & conv & 4$\times$4 & 2 & 64\\
		\cline{2-5} \cline{8-11} \cline{13-17}
		& conv2 & 3$\times$3 & 1 & 64 & && up-conv2 & 3$\times$3 & 1 & 512 && \multicolumn{1}{|c|}{C2} & conv & 4$\times$4 & 2 & 64\\
        \cline{1-5} \cline{7-11} \cline{13-17}
		\multirow{2}{*}{C2} & conv1 & 2$\times$2 & 2 &128 && \multirow{2}{*}{DC2} & up-conv1 & 2$\times$2 & 2 &256 && \multicolumn{1}{|c|}{C3} & conv & 4$\times$4 & 2 & 128\\
		\cline{2-5} \cline{8-11} \cline{13-17}
		& conv2 & 3$\times$3 & 1 & 128 & && up-conv2 & 3$\times$3 & 1 & 256 && \multicolumn{1}{|c|}{C4} & conv & 4$\times$4 & 2 & 128\\
		\cline{1-5} \cline{7-11} \cline{13-17}
		\multirow{2}{*}{C3} & conv1 & 2$\times$2 & 2 &256 && \multirow{2}{*}{DC3} & up-conv1 & 2$\times$2 & 2 &128 && \multicolumn{1}{|c|}{C5} & conv & 4$\times$4 & 2 & 256\\
		\cline{2-5} \cline{8-11} \cline{13-17}
		& conv2 & 3$\times$3 & 1 & 256 & && up-conv2 & 3$\times$3 & 1 & 128 && \multicolumn{1}{|c|}{C6} & conv & 4$\times$4 & 2 & 256\\
        \cline{1-5} \cline{7-11} \cline{13-17}
		\multirow{2}{*}{C4} & conv1 & 2$\times$2 & 2 &512 && \multirow{2}{*}{DC4} & up-conv1 & 2$\times$2 & 2 &64 && \multicolumn{1}{|c|}{C7} & conv & 3$\times$3 & 1 & 1\\
		\cline{2-5} \cline{8-11} \cline{13-17}
		& conv2 & 3$\times$3 & 1 & 512 & && up-conv2 & 3$\times$3 & 1 & 64\\
		\cline{1-5} \cline{7-11} 
		C5 & conv1 & 2$\times$2 & 2 &1024 && DC5 & up-conv1 & 3$\times$3 & 1 &1\\
		\cline{1-11}
	\end{tabular}
	\begin{tablenotes}
		\footnotesize
		\item conv: convolution. up-conv: up-convolution.
	\end{tablenotes}
\vspace{-0.05in}
\end{table*}

\subsubsection{Discriminator}
The architecture of the discriminator is a classical convolutional neural network (CNN). It has seven composite blocks, and each of the first six blocks consists of a convolutional layer followed by a batch-normalization layer and a leaky ReLU layer. The last block consists of a convolutional layer, a batch-normalization layer, and a sigmoid layer. Similar to the parameter choice of the SC-DAE, we use small kernels for the discriminator. Details of the kernels are in Table \ref{table:parameters}. 

The discriminator takes a two-channel map as input and outputs a binary classification score. Specifically, the SC-DAE generated enhanced latent fingerprints and the ground-truth skeleton map are respectively concatenated with the ground-truth orientation field to form two types of two-channel inputs to the discriminator. The discriminator tries to distinguish them and thus can force the SC-DAE generated enhanced latent fingerprint indistinguishable from the ground truth in terms of the fingerprint skeleton map and the FOMFE-based orientation field. This way, it enables the SC-DAE to have an ability of deep semantic understanding, and thus to learn to restore the corrupted ridge structure of the latent fingerprint in addition to the de-noising.

\subsubsection{Gaussian-based Minutia Weight Map}
\begin{figure}[htbp]
 	\centering
 	\includegraphics[width=\linewidth]{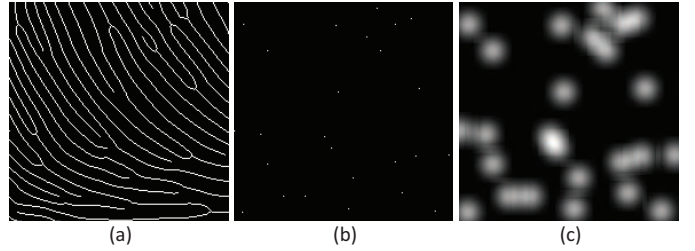}
 	\caption{Illustration of the proposed Gaussian-based minutia weight map. (a) a fingerprint skeleton map; (b) the minutia map $M$ of (a); and (c) the Gaussian-based minutia weight map of (a).}
 	\label{fig:weightmap}
 \end{figure}

 To force the SC-DAE optimize minutia information, we propose a Gaussian-based minutia weight map $w$ which is defined as:
\begin{equation}
w(x,y) = 
\begin{cases}
w'(x,y),  & \mbox{if } w'(x,y) \neq 0, \\
w_0 , & \mbox{otherwise},
\end{cases}
\end{equation}
with
\begin{equation}
w'(x,y) = \frac{\sum_{u=-r}^{r}\sum_{v=-r}^{r}w_g(u,v)\cdot M(x+u,y+v)}{\sum_{u=-r}^{r}\sum_{v=-r}^{r}w_g(u,v)},
\end{equation}

\begin{equation}
w_g(u,v) = \frac{1}{2\pi \sigma^2}e^{-\frac{u^2 +v^2}{2 \sigma^2}},
\end{equation}
and
\begin{equation}
w_0 = \frac{w_g(r,r)}{\sum_{u=-r}^{r}\sum_{v=-r}^{r}w_g(u,v)},
\end{equation}
 where ($x, y$) are coordinates of each pixel, ($u, v$) are coordinates of pixels in a local window centered at ($x,y$), $r$ is half the size of the local window, $M$ is the minutia map whose value is 1 at minutia and 0 otherwise (as shown in Fig. \ref{fig:weightmap}(b)), and $\sigma$ is the standard deviation of Gaussian. In the experiment, $\sigma$ is set to be $8$ and $r$ is set to be $17$. An example of the proposed weight map is illustrated in Fig. \ref{fig:weightmap}.

\subsubsection{Loss Functions}
 The FingerGAN has two losses: 1) an adversarial loss which is used to jointly train the discriminator and the SC-DAE, and force the SC-DAE to generate enhanced latent fingerprint indistinguishable from the ground-truth in terms of fingerprint skeleton map and FOMFE-based orientation field; and 2) a reconstruction loss which is used to further force the SC-DAE to generate enhanced latent fingerprint in the context of optimizing minutiae information.

Denote the training latent fingerprint as $l$ and its domain as $\mathcal{L}$, the SC-DAE as $G$, the generated enhanced latent fingerprint as $G(l)$, the ground-truth skeleton map as $g$ and its domain as $\mathcal{G}$, the ground-truth FOMFE-based orientation field as $g_F$ and its domain as $\mathcal{G_F}$ \cite{39wang2007fingerprint}, and the discriminator as $D$. According to the loss function in Eq.(\ref{eq:gan}), the adversarial loss $L_a$ is formulated as:
\begin{equation}
\begin{aligned}
\min_{G} \max_{D} L_a(G,D) & = \mathbb{E}_{g \in \mathcal{G}, g_F \in \mathcal{G_F}}[log(D(g, g_F))]  \\
& + \mathbb{E}_{l \in \mathcal{L}, g_F \in \mathcal{G_F}}[log(1-D(G(l), g_F))].
\end{aligned}
\end{equation}
We use the L1 loss as the reconstruction loss $L_r$, and thus it is formulated as:
\begin{equation}
\label{eq:reloss}
L_r(G) = \mathbb{E}_{l \in \mathcal{L}}[|| w \odot(g - G(l))||_1],
\end{equation}
where $\odot$ denotes the element-wise multiplication. Overall, the total loss function is formulated as: 
\begin{equation}
\min_{G} \max_{D} L = L_{a} + \eta L_{r},
\end{equation}
where $\eta$ is a parameter that weights the contributions of the reconstruction loss and the adversarial loss. It is empirically set to be $0.001$ in the experiments.

\subsection{Proposed Training Data Generation}
\label{subsec:dataprepare}
Applying deep learning to latent fingerprint applications is challenging because the current public databases are either short of the correspondence between latent fingerprints and their true mates or lack of quantity. In this paper, we propose an effective procedure to generate the training data.

\subsubsection{Overview}
\begin{figure*}[htbp]
	\centering
	\includegraphics[width=0.9\linewidth]{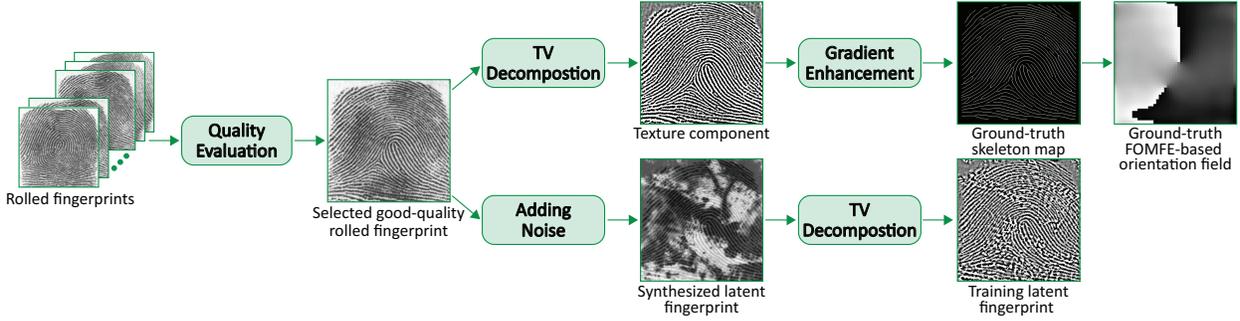}
	\caption{Schematic diagram of the proposed training data generation.}
	\label{fig:datapre}
\end{figure*}
Fig. \ref{fig:datapre} illustrates the process of the proposed training data generation. Firstly, a quality evaluation is performed on rolled fingerprints to select those good-quality ones. This is important because it ensures obtaining reliable ground-truth labels to provide meaningful supervision for the training. Then, the TV decomposition \cite{RN761} is applied to those selected good-quality fingerprints to obtain their texture components which are subsequently enhanced and thinned to generate ground-truth skeleton maps. Also, ground-truth FOMFE-based orientation fields are calculated based on the ground-truth skeleton maps using the method in \cite{39wang2007fingerprint}. Meanwhile, noise is added to those selected good-quality fingerprints to obtain synthesized latent fingerprints which are subsequently decomposed by the TV decomposition to obtain their texture components to be used as the training latent fingerprints. In the experiment, the rolled fingerprints are from the database NIST SD14 \cite{33NISTSD14}. The quality evaluation is achieved using the method in \cite{34MINDTCT} due to its effectiveness. The enhancement is achieved using the gradient-based method in \cite{4chikkerur2007fingerprint} due to the good quality of those selected fingerprints. The noise is added by the proposed latent fingerprint synthesization method which is described as follows.

\subsubsection{Latent Fingerprint Synthesization}
For better noise simulation, we propose adding complex and realistic noise instead of simple lines or characters noise adopted in previous works \cite{18cao2015latent,20li2018deep}. This helps provide abundant training data that are better mimicking real latent fingerprint cases and is important for the SC-DAE to learn more effective representations of fingerprints from tough situations. 

\begin{figure}[htbp]
	\centering
	\includegraphics[width=\linewidth]{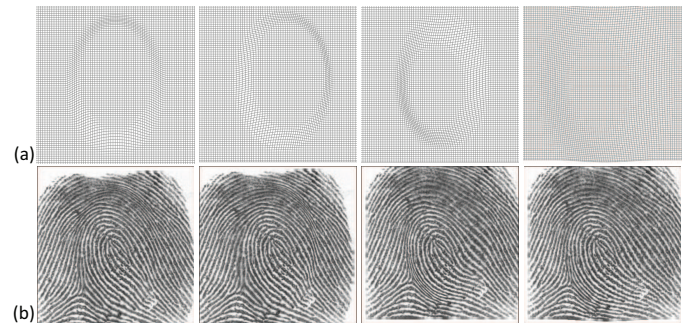}
	\caption{ (a) various plastic distortions and (b) corresponding distorted fingerprints based on the same rolled fingerprint.}
	\label{fig:distortion}
\end{figure}

Given a selected rolled fingerprint $b$, firstly, a plastic distortion \cite{49RN403} is added by the following equation:
\begin{equation}
b'_i = b_i +\Delta (b_i)\cdot g(h(b_i),k),
\end{equation}
where $b_i = [x_{b_i}, y_{b_i}]^T$ is a point in $b$ and $b'_i $ is its distorted point. $k$ is the skin plasticity coefficient. $\Delta (b_i)$ is the torsion and traction amount computed on the basis of a rotation angle $\theta$ and a displacement vector $e = [e_x, e_y]^T$, and is given by
\begin{equation}
\Delta (b_i) = (R_{\theta}\cdot(b_i-o_r)+o_r+e)-b_i,
\end{equation}
with
\begin{equation}
R_{\theta} =  \begin{bmatrix}
cos \theta & sin\theta \\ 
-sin\theta & cos\theta 
\end{bmatrix},
\end{equation}
where $o_r$ is the center of rotation. $g(h(b_i),k)$ is the gradual transition defined as:
\begin{equation}
g(h(b_i),k) = \left\{\begin{array}{@{}ll}
0 & h(b_i) < 0\\ 
\frac{1}{2}(1 - cos(\frac{\pi \cdot h(b_i)}{k})) & 0< h(b_i) < k ,\\ 
1 & otherwise
\end{array}\right.
\end{equation} 
where function $h(b_i)$ returns a measure proportional to the distance between the point and the border of an ellipse centered at $o_e$ with semi-axes $s_x$ and $s_y$, and is formulated as:
\begin{equation}
h(b_i) = \sqrt{(b_i-o_e)^TA^{-1}(b_i-o_e)-1},
\end{equation}
with
\begin{equation}
A =  \begin{bmatrix}
s^2_x & 0 \\ 
0 & s^2_y
\end{bmatrix}.
\end{equation}
In the experiments, to generate reasonable distortions, the ranges of values for parameters $k$, $\theta$, $e$, and $A$ are empirically set to $[0.5, 2]$, $[0, 5]$, $[-15, 15],$ and $\{[0.2s, 0.6s]$ (for $s_x$), $[s_x, 2s_x]$ (for $s_y$)$\}$, respectively, where $s$ is half of the size of the fingerprint image width. $o_r$ and $o_e$ are both set to be the center of the fingerprint image. Fig. \ref{fig:distortion} shows various plastic distortions and their corresponding distorted fingerprints based on the same rolled fingerprint. 

\begin{figure}[h]
	\centering
	\includegraphics[width=\linewidth]{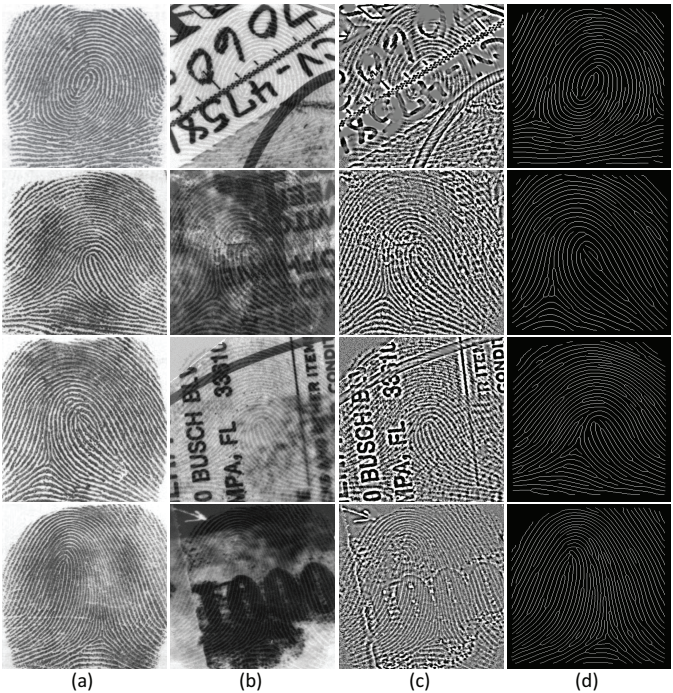}
	\caption{Examples illustrating the proposed training data generation by (a) selected good-quality rolled fingerprints, (b) synthesized latent fingerprints of (a), (c) training latent fingerprints (TV decomposed textures) of (a), and (d) ground-truth skeleton maps of (a).}
	\label{fig:latent}
\end{figure}

Then, speckle noise is added to the distorted fingerprint $b'$ by the equation $b'' = b'+n*b'$, where $n$ is uniformly distributed random noise with mean $0$ and variance set to $(0, 0.02)$. Finally, a latent fingerprint $c$ is synthesized by fusing $b''$ and a realistic noise image $d$ according to the equation: 
\begin{equation}
c = (1-\lambda ) b'' + \lambda d,
\end{equation} 
where $d$ is randomly cropped from the background regions of latent fingerprints in the NIST SD27 database \cite{33NISTSD14}. $\lambda $ is a weight that measures the intensity degree of the realistic noise image. In the experiments, its value ranges from $0.2$ to $0.8$. Fig. \ref{fig:latent} shows some synthesized latent fingerprints and their training latent fingerprints and ground-truth skeleton maps generated by the proposed latent fingerprint synthesization method.

\section{Experimental Results}
\label{sec:experimentalresults}
In this section, we evaluate our proposed method. Database and implementation details are firstly introduced in Section \ref{subsec:implementation}. Experimental results are then presented in Section \ref{subsec:visual} and \ref{subsec:quantitative}. Finally, the proposed method is analyzed and discussed in Section \ref{subsec:analysis}.

\subsection{Database and Implementation Details}
\label{subsec:implementation}
\subsubsection{Database} 
\textbf{Training Database} The database NIST SD14 \cite{33NISTSD14} is used to generate the training data, which consists of $54,000$ rolled fingerprint. According to the described method in Section \ref{subsec:dataprepare}, a total of $13,000$ good-quality fingerprints are selected from them. For each of the selected fingerprints, $10$ latent fingerprints are synthesized, and thus the final training database consists of a total of $130,000$ training latent fingerprints and $13,000$ corresponding ground-truth skeleton maps. 

\textbf{Testing Databases} Two challenging latent fingerprint databases NIST SD27 \cite{41NIST} and IIIT-Delhi MOLF \cite{43sankaran2015multisensor} are used to evaluate the performance of the proposed method. Database NIST SD27 is provided by the National Institute of Standards and Technology in collaboration with the FBI. It contains $258$ latent fingerprint images collected from crime scenes. Latent fingerprints in this database contain complex noises and degradations of various types and levels, and therefore is a rigorous benchmark for evaluating the performance of the proposed method. Database IIIT-Delhi MOLF is provided by Sankaran et al. and is widely used in latent fingerprint tasks in recent years. It contains $4,400$ latent fingerprints and three sets of live-scan fingerprints obtained by different acquisition sensors of 'Crossmatch', 'Secugen', and 'Lumidigm'. Each set has $4,000$ live-scan fingerprints and can be used as a reference database for latent fingerprint identification. These three reference databases are denoted as 'C', 'S', and 'L', respectively. The resolution of images in these two databases are $500$ppi.

\subsubsection{Implementation Details} 
\label{subsubsec:implementationdetails}
\textbf{Enhancement} Details of the architecture of the FingerGAN are provided in Fig. \ref{fig:aa} and Table \ref{table:parameters}. It was implemented in PyTorch and its optimizations are solved by SGD solver Adam \cite{42kingma2014adam} with a learning rate of $0.001$. During the training, $192 \times 192$ patches are used to train the FingerGAN. During the testing, for a latent fingerprint to be enhanced, a sliding window of size $192 \times 192$ with a step size of $8$ was adopted to generate the enhanced latent fingerprint using the trained SC-DAE. Implementation codes will be available for non-commercial purposes from \url{https://github.com/HubYZ/LatentEnhancement}.

\textbf{Identification} Enhanced latent fingerprint identification experiments are conducted to quantitatively evaluate the performance of the proposed method. For experiments conducted on the NIST SD27 database, the manually marked regions of interest provided in \cite{8feng2013orientation} are used. Also, to make the identification more challenging, the reference fingerprint database is extended by adding rolled fingerprints from the  NIST SD14 database. This is reasonable because the NIST SD14 database has been only used for the enhancement training and has not been used in any way for the identification task. Therefore, each enhanced latent fingerprint is compared with a total of $27,258$ rolled fingerprints for the identification. For experiments conducted on the IIIT-Delhi MOLF database, each enhanced latent fingerprint is compared with the first and second fingerprint samples of each subject for each of the three reference databases according to the test protocol established by Sankaran et al. \cite{43sankaran2015multisensor}. The commercial software Neurotechnology VeriFinger SDK12.1 \footnote{https://www.neurotechnology.com/release-notes-megamatcher-120.html\#2021-03-26.} is used for the identification. Cumulative Match Characteristic (CMC) curve is employed to evaluate the performance of the latent fingerprint identification.

\subsection{Minutia Recovery Accuracy}
\label{subsec:visual}
\subsubsection{Quantitative Evaluation}
\begin{figure*}[b]
	\centering
	\includegraphics[width=0.8\linewidth]{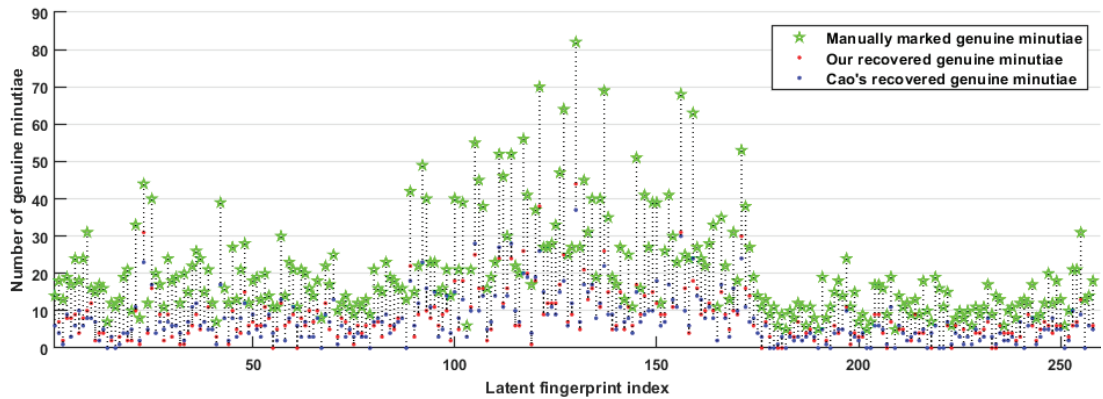}
	\caption{Numbers of recovered genuine minutiae extracted from our and Cao’s enhanced latent fingerprints, compared with the numbers of manually marked genuine minutiae for each of the 258 latent fingerprints in the NSIT SD27 database.}
	\label{fig:genuineminutiae}
\end{figure*}
\begin{figure*}[b]
	\centering
	\includegraphics[width=0.8\linewidth]{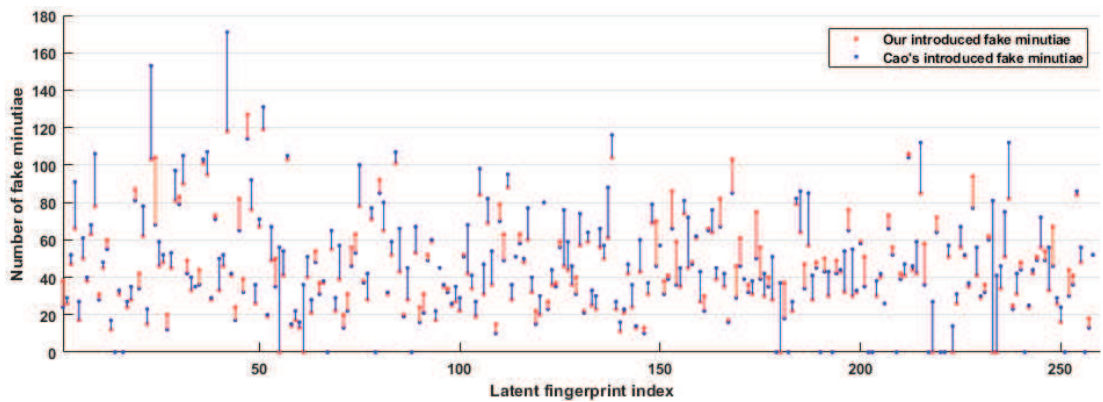}
	\caption{Comparison of numbers of introduced fake minutiae in our and Cao’s enhanced latent fingerprints for each of the 258 latent fingerprints in the NSIT SD27 database.}
	\label{fig:fakeminutiae}
\end{figure*}
To evaluate the performance of our proposed method on latent fingerprint enhancement, we investigate its minutia recovery accuracy and compare it with that of the state-of-the-art method \cite{RN930} (Cao's method). This experiment is conducted on the NIST SD27 database because it provides manually marked minutiae which can be used as genuine minutiae. We enhance the latent fingerprints by our and Cao's methods, respectively, and then extract minutiae from our and Cao's enhanced latent fingerprints using the VeriFinger12.1. We compare these extracted minutiae with the genuine minutiae and define the recovered genuine minutiae as those extracted minutiae with both correct location, orientation, and minutia type in accordance with the genuine minutiae. All the other extracted minutiae are defined as introduced fake minutiae. Table \ref{table:minutiae1} shows the comparison results of Cao's and our methods. As can be seen, our method recovers more genuine minutiae meanwhile introduces fewer fake minutiae than Cao's method. Detailed comparisons on each of the 258 enhanced latent fingerprints are provided in Fig. \ref{fig:genuineminutiae} and Fig. \ref{fig:fakeminutiae}. As can be seen, there are $117$ latent fingerprints where our enhanced latent fingerprints recover more genuine minutiae than Cao's enhanced latent fingerprints, while there are $94$ latent fingerprints where Cao's enhanced latent fingerprints recover more genuine minutiae than our enhanced latent fingerprints. Furthermore, there are $149$ latent fingerprints where our enhanced latent fingerprints introduce fewer fake minutiae than Cao's enhanced latent fingerprints, while there are $83$ latent fingerprints where Cao's enhanced latent fingerprints introduce fewer fake minutiae than our enhanced latent fingerprints. These results demonstrate the superiority of our method in terms of minutia recovery accuracy, and support our claim that the FingerGAN can perform latent fingerprint enhancement in the context of directly optimizing minutia information. 

\begin{table}[htbp]
	\renewcommand{\arraystretch}{1.6}
	\caption{Comparison of minutia recovery accuracy of different methods in terms of overall numbers of recovered genuine minutiae and introduced fake minutiae for the 258 latent fingerprints in the NIST SD27 database.}
	\label{table:minutiae1}
	\centering
	\begin{tabular}{|c|c|c|}
		\hline
		\diagbox{Metrics}{Methods}& ~~~~Cao's~~~~& ~~~~Ours~~~~ \\
		\hline
		~~Recovered genuine minutiae~~& 1,887& 1,982\\
		\hline
		Introduced fake minutiae & 12,235& 11,152 \\
		\hline
	\end{tabular}
\end{table}

\subsubsection{Visual Inspection}
\begin{figure*}[htbp]
	\centering
	\includegraphics[width=0.8\linewidth]{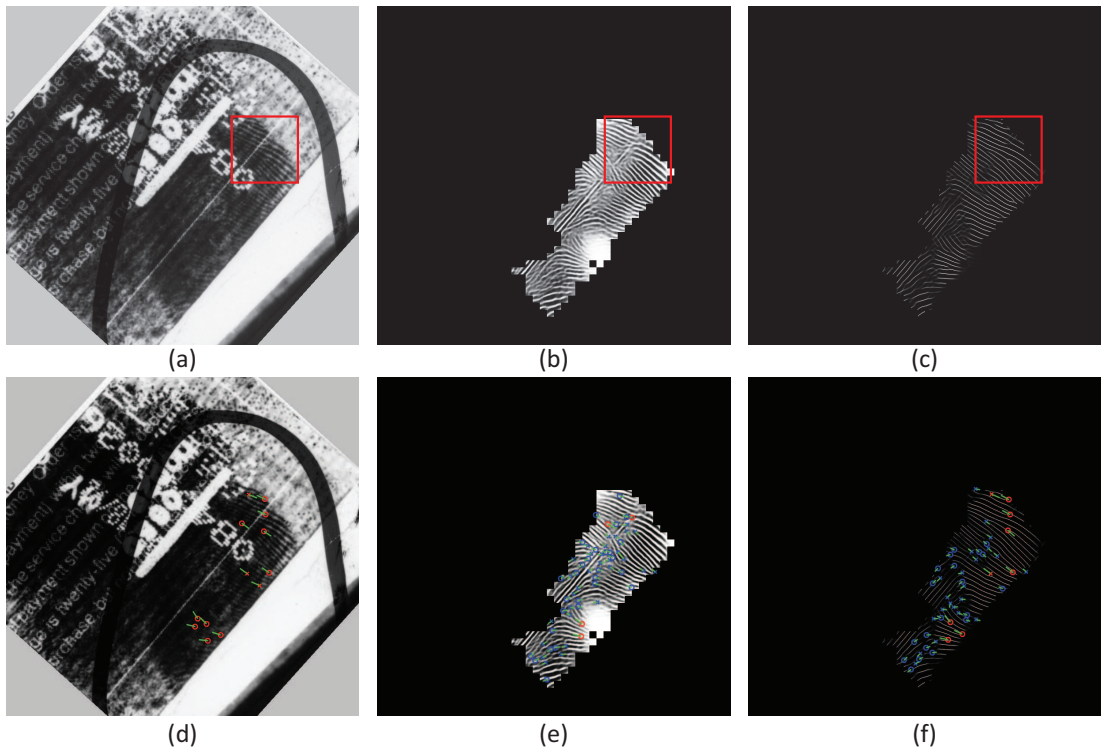}
	\caption{An illustrative example comparing the results of our method with Cao's method. (a) An ugly latent fingerprint U250 from the NIST SD27 database, (b) the enhanced latent fingerprint by Cao's method, which is recognized as rank-131 with a matching score of $6$, (c) the enhanced latent fingerprint by our proposed method, which is recognized as rank-1 with a matching score of $38$, (d) the manually marked minutiae on latent fingerprint (a), (e) extracted minutiae on Cao's enhanced latent fingerprint (b), and (f) extracted minutiae on our enhanced latent fingerprint (c). Genuine minutiae are labeled as red circle or cross with their orientations labeled as green line. Fake minutiae are labeled as blue circle or cross with their orientations labeled as green line.}
	\label{fig:visualexample}
\end{figure*}
We provide an illustrative example in Fig. \ref{fig:visualexample} to compare our method with Cao's method, which shows the enhanced latent fingerprints and the recovered minutiae against manually marked minutiae. By observing and comparing the Fig. \ref{fig:visualexample}(a), (b), and (c), especially the upper right areas (red rectangles) of the fingerprints, we can observe that our enhanced latent fingerprint (c) gets better ridge/valley structures than Cao's enhanced latent fingerprint (b). The superiority of our method can also be proved by observing the recovered minutiae from the Fig. \ref{fig:visualexample}(d), (e), and (f). As can be seen, a total of $13$ minutiae are manually marked in the latent fingerprint (d), only four recovered genuine minutiae are extracted from Cao's enhanced latent fingerprint (e), however, nine recovered genuine minutiae are extracted from our enhanced latent fingerprint (f). Also, Cao's enhanced latent fingerprint introduces $59$ fake minutiae, however, our enhanced latent fingerprint introduces $42$ fake minutiae.

\subsection{Identification Performance} 
\label{subsec:quantitative}
\subsubsection{Evaluation on Database NIST SD27}
\begin{figure}[h]
	\centering
	\includegraphics[width=0.9\linewidth]{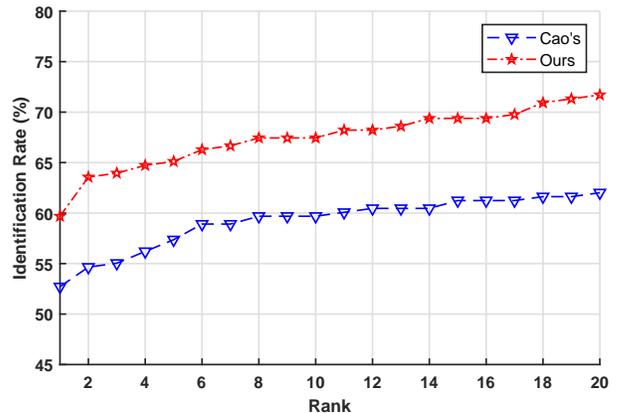}
	\caption{Comparison of CMC curves achieved using the enhanced latent fingerprints generated by our and Cao's methods.}
	\label{fig:ROCcomparison_sd27}
\end{figure}

To comprehensively evaluate our proposed method, we conduct fingerprint identification experiments using the enhanced latent fingerprints. We compare the performance of our method with that of Cao's method. Fig. \ref{fig:ROCcomparison_sd27} shows the comparison of CMC curves achieved using the enhanced latent fingerprints generated by our and Cao's methods. As can be seen, the rank-1 accuracy of our method is $59.69\%$, which is significantly better than that of Cao's method ($52.71\%$). Our method achieves an improvement of $13.24\%$, which demonstrates its superiority in latent fingerprint enhancement.

\subsubsection{Evaluation on Database IIIT-Delhi MOLF}
\begin{figure}[htbp]
	\centering
	\includegraphics[width=0.9\linewidth]{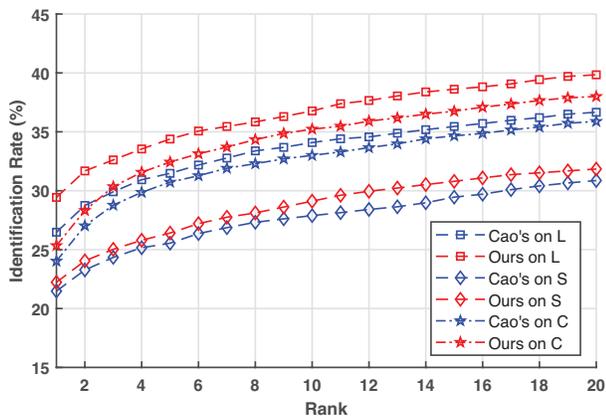}
	\caption{Comparison of CMC curves achieved using the enhanced latent fingerprints generated by our and Cao's methods over the three reference databases 'C', 'S', and 'L', respectively.}
	\label{fig:ROCcomparison_IIIT}
\end{figure}

We also compare the identification performance of our method with that of Cao's method on the IIIT-Delhi MOLF database. Fig. \ref{fig:ROCcomparison_IIIT} shows the comparison of CMC curves achieved using the enhanced latent fingerprints generated by our and Cao's methods over the three reference databases 'C', 'S', and 'L', respectively. As can be seen, the rank-1 accuracies of our method evaluated using the 'C', 'S', and 'L' reference databases are $25.34\%$, $22.23\%$, and $29.43\%$, respectively, which are consistently better than those of Cao's method ($24.02\%$, $21.47\%$, and $26.47\%$). Our method obtains improvements of $5.5\%$, $3.54\%$, and $11.18\%$, respectively. These results also demonstrate the superiority of our method in latent fingerprint enhancement.

\subsection{Ablation Study}
\label{subsec:analysis}
To further analyze our method and justify the effectiveness of the design of the FingerGAN, we conduct the following ablation studies. These experiments are conducted on the NIST SD27 database using a reference database consisting of $258$ corresponding rolled fingerprints of the NIST SD27 database. All the other experimental settings are the same as those described in Section \ref{subsubsec:implementationdetails}, except stated otherwise. 

\subsubsection{The Advantage of Embedding the SC-DAE in a GAN}
\begin{figure}[htbp]
	\centering
	\includegraphics[width=0.8\linewidth]{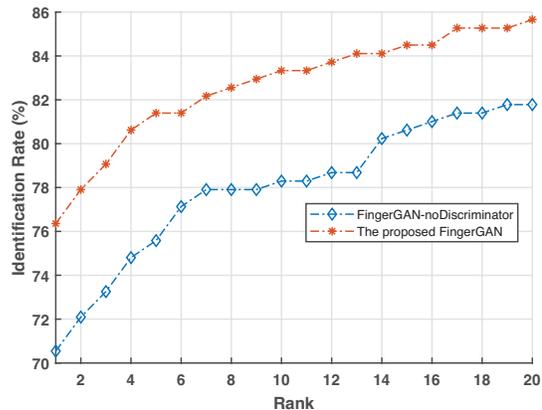}
	\caption{Comparison of CMC curves achieved using the proposed FingerGAN and the FingerGAN-noDiscriminator.}
	\label{fig:ablation1}
\end{figure}

To demonstrate the effectiveness of embedding the SC-DAE in a GAN, we conducted the following ablation study. We use only the proposed SC-DAE for latent fingerprint enhancement without using the discriminator, and name this method as FingerGAN-noDiscriminator. Specifically, we train the SC-DAE using only the reconstruction loss in Eq.(\ref{eq:reloss}) with only the fingerprint skeleton maps as the ground-truths. Fig. \ref{fig:ablation1} compares the CMC curves achieved using the proposed FingerGAN and the FingerGAN-noDiscriminator. As can be seen, the rank-1 accuracy achieved using the FingerGAN ($76.36\%$) is significantly higher than that achieved using the FingerGAN-noDiscriminator ($70.54\%$). This demonstrates the effectiveness of embedding the SC-DAE in a GAN and supports our claim that the proposed FingerGAN can force its generated enhanced latent fingerprints indistinguishable from the ground-truths. 

\subsubsection{The Advantage of Using the Skeleton Map}
\begin{figure}[htbp]
	\centering
	\includegraphics[width=0.8\linewidth]{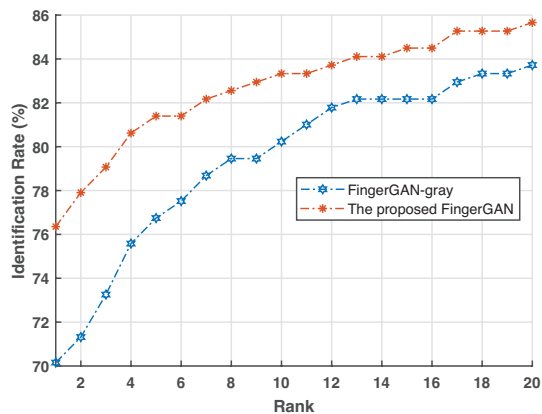}
	\caption{Comparison of CMC curves achieved using the proposed FingerGAN and the FingerGAN-gray.}
	\label{fig:ablation2}
\end{figure}

To demonstrate the effectiveness of using the fingerprint skeleton map as a ground-truth, we conducted the following ablation study. We use fingerprint gray images instead of fingerprint skeleton maps as the ground-truths to train the FingerGAN, and name this method as FingerGAN-gray. Fig. \ref{fig:ablation2} compares the CMC curves achieved using the proposed FingerGAN and the FingerGAN-gray. As can be seen, the rank-1 accuracy achieved using the proposed FingerGAN ($76.36\%$) is significantly higher than that achieved using the FingerGAN-gray ($70.15\%$). This demonstrates the effectiveness of using the fingerprint skeleton map as a ground-truth.

\subsubsection{The Advantage of the Gaussian Minutia Weight}
\begin{figure}[htbp]
	\centering
	\includegraphics[width=0.8\linewidth]{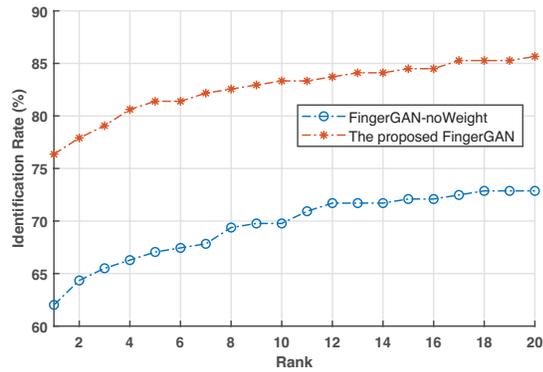}
	\caption{Comparison of CMC curves achieved using the proposed FingerGAN and the FingerGAN-noWeight.}
	\label{fig:ablation3}
\end{figure}

To demonstrate the effectiveness of using the Gaussian-based minutia weight map, we conducted the following ablation study. We train the FingerGAN using a loss function without the Gaussian-based minutia weight map, and name this method as FingerGAN-noWeight. That is, we remove $w$ in the Eq.(\ref{eq:reloss}) and make the reconstruction loss as $L_r(G) = \mathbb{E}_{l \in \mathcal{L}}[||g - G(l)||_1]$ to train the FingerGAN. Fig. \ref{fig:ablation3} compares the CMC curves achieved using the proposed FingerGAN and the FingerGAN-noWeight. As can be seen, the rank-1 accuracy achieved using the proposed FingerGAN ($76.36\%$) is significantly higher than that achieved using the FingerGAN-noWeight ($62.02\%$). This demonstrates the effectiveness of using the Gaussian-based minutia weight map and supports our claim that we offer a FingerGAN that can perform latent fingerprint enhancement in the context of optimizing minutia information.

\section{Conclusions}
\label{sec:conclusion}
This paper proposed a FingerGAN for latent fingerprint enhancement, which formulates the latent fingerprint enhancement as a constrained fingerprint generation problem. It can enforce its generated enhanced latent fingerprint indistinguishable from the corresponding ground-truth instance in terms of the fingerprint skeleton map weighted by minutia locations and the orientation field regularized by the FOMFE model. Because minutia is the primary feature for recognition and minutia can be retrieved directly from the fingerprint skeleton map, we offer a holistic framework which can perform latent fingerprint enhancement in the context of directly optimizing minutia information. This will help improve latent fingerprint identification performance significantly. Experimental results on two public latent fingerprint databases demonstrate that our method outperforms the state of the arts significantly.

\ifCLASSOPTIONcompsoc

\section*{Acknowledgments}
\else

\section*{Acknowledgment}
\fi
This research was undertaken with the assistance of resources and services provided by the National Computational Infrastructure (NCI) organization, which is supported by the Australian Government. This project is partially supported by ARC Discovery Grants (DP190103660 and DP200103207) and ARC Linkage Grant (LP180100663).

\ifCLASSOPTIONcaptionsoff
\newpage
\fi

\bibliographystyle{IEEEtran}
\bibliography{reference}

\begin{thebibliography}{10}
\providecommand{\url}[1]{#1}
\csname url@samestyle\endcsname
\providecommand{\newblock}{\relax}
\providecommand{\bibinfo}[2]{#2}
\providecommand{\BIBentrySTDinterwordspacing}{\spaceskip=0pt\relax}
\providecommand{\BIBentryALTinterwordstretchfactor}{4}
\providecommand{\BIBentryALTinterwordspacing}{\spaceskip=\fontdimen2\font plus
\BIBentryALTinterwordstretchfactor\fontdimen3\font minus
  \fontdimen4\font\relax}
\providecommand{\BIBforeignlanguage}[2]{{%
\expandafter\ifx\csname l@#1\endcsname\relax
\typeout{** WARNING: IEEEtran.bst: No hyphenation pattern has been}%
\typeout{** loaded for the language `#1'. Using the pattern for}%
\typeout{** the default language instead.}%
\else
\language=\csname l@#1\endcsname
\fi
#2}}
\providecommand{\BIBdecl}{\relax}
\BIBdecl

\bibitem{1HawthorneFingerprints}
M.~Hawthorne, \emph{Fingerprints: Analysis and Understanding}.\hskip 1em plus
  0.5em minus 0.4em\relax CRC Press, 2008.

\bibitem{2MaltoniHandbook}
D.~Maltoni, D.~Maio, A.~K. Jain, and S.~Prabjakar, \emph{Handbook of
  Fingerprint Recognition second ed}.\hskip 1em plus 0.5em minus 0.4em\relax
  Springer-Verlag, 2009.

\bibitem{wallace2004detection}
C.~Wallace-Kunkel, C.~Roux, C.~Lennard, and M.~Stoilovic, ``The detection and
  enhancement of latent fingermarks on porous surfaces-a survey,''
  \emph{Journal of Forensic Identification}, vol.~54, no.~6, p. 687, 2004.

\bibitem{sankaran2014latent}
A.~Sankaran, M.~Vatsa, and R.~Singh, ``Latent fingerprint matching: A survey,''
  \emph{IEEE Access}, vol.~2, pp. 982--1004, 2014.

\bibitem{malwade2015survey}
A.~V. Malwade, R.~D. Raut, and V.~Thakare, ``A survey on fingerprint
  enhancement techniques using filtering approach,'' \emph{International
  Journal of Electronics, Communication and Soft Computing Science \&
  Engineering (IJECSCSE)}, p. 372, 2015.

\bibitem{schuch2018survey}
P.~Schuch, S.~Schulz, and C.~Busch, ``Survey on the impact of fingerprint image
  enhancement,'' \emph{IET Biometrics}, vol.~7, no.~2, pp. 102--115, 2018.

\bibitem{abebe2020latent}
B.~Abebe, H.~C.~A. Murthy, E.~Amare~Zereffa, and Y.~Dessie, ``Latent
  fingerprint enhancement techniques: A review,'' \emph{Journal of Chemical
  Reviews}, vol.~2, no.~1, pp. 40--56, 2020.

\bibitem{3cappelli2009semi}
R.~Cappelli, D.~Maio, and D.~Maltoni, ``Semi-automatic enhancement of very low
  quality fingerprints,'' in \emph{International Symposium on Image and Signal
  Processing and Analysis}, 2009, pp. 678--683.

\bibitem{4chikkerur2007fingerprint}
S.~Chikkerur, A.~N. Cartwright, and V.~Govindaraju, ``Fingerprint enhancement
  using {STFT} analysis,'' \emph{Pattern recognition}, vol.~40, no.~1, pp.
  198--211, 2007.

\bibitem{6yoon2010latent}
S.~Yoon, J.~Feng, and A.~K. Jain, ``On latent fingerprint enhancement,'' in
  \emph{Biometric Technology for Human Identification VII}, vol. 7667, 2010, p.
  766707.

\bibitem{7yoon2011latent}
S.~Yoon, J.~Feng, and A.~K. Jain, ``Latent fingerprint enhancement via robust
  orientation field estimation,'' in \emph{International joint conference on
  biometrics}, 2011, pp. 1--8.

\bibitem{8feng2013orientation}
J.~Feng, J.~Zhou, and A.~K. Jain, ``Orientation field estimation for latent
  fingerprint enhancement,'' \emph{IEEE Trans. on Pattern Analysis and Machine
  Intelligence}, vol.~35, no.~4, pp. 925--940, 2013.

\bibitem{9yang2014localized}
X.~Yang, J.~Feng, and J.~Zhou, ``Localized dictionaries based orientation field
  estimation for latent fingerprints,'' \emph{IEEE Trans. on PAMI}, vol.~36,
  no.~5, pp. 955--969, 2014.

\bibitem{10zhang2012latent}
J.~Zhang, R.~Lai, and C.-C.~J. Kuo, ``Latent fingerprint segmentation with
  adaptive total variation model,'' in \emph{International Conference on
  Biometrics}, 2012, pp. 189--195.

\bibitem{11zhang2013adaptive}
J.~Zhang, R.~Lai, and C.-C.~J. Kuo, ``Adaptive directional total-variation
  model for latent fingerprint segmentation,'' \emph{IEEE Trans. on IFS},
  vol.~8, no.~8, pp. 1261--1273, 2013.

\bibitem{12cao2014segmentation}
K.~Cao, E.~Liu, and A.~K. Jain, ``Segmentation and enhancement of latent
  fingerprints: A coarse to fine ridgestructure dictionary,'' \emph{IEEE Trans.
  on PAMI}, vol.~36, no.~9, pp. 1847--1859, 2014.

\bibitem{13liu2015latent}
M.~Liu, X.~Chen, and X.~Wang, ``Latent fingerprint enhancement via multi-scale
  patch based sparse representation,'' \emph{IEEE Trans. on IFS}, vol.~10,
  no.~1, pp. 6--15, 2015.

\bibitem{tang2017fingernet}
Y.~Tang, F.~Gao, J.~Feng, and Y.~Liu, ``Fingernet: An unified deep network for
  fingerprint minutiae extraction,'' in \emph{2017 IEEE International Joint
  Conference on Biometrics (IJCB)}.\hskip 1em plus 0.5em minus 0.4em\relax
  IEEE, 2017, pp. 108--116.

\bibitem{nguyen2018robust}
D.-L. Nguyen, K.~Cao, and A.~K. Jain, ``Robust minutiae extractor: Integrating
  deep networks and fingerprint domain knowledge,'' in \emph{2018 International
  Conference on Biometrics (ICB)}.\hskip 1em plus 0.5em minus 0.4em\relax IEEE,
  2018, pp. 9--16.

\bibitem{18cao2015latent}
K.~Cao and A.~K. Jain, ``Latent orientation field estimation via convolutional
  neural network,'' in \emph{International Conference on Biometrics}, 2015, pp.
  349--356.

\bibitem{19svoboda2017generative}
J.~Svoboda, F.~Monti, and M.~M. Bronstein, ``Generative convolutional networks
  for latent fingerprint reconstruction,'' in \emph{IEEE International Joint
  Conference on Biometrics}, 2017, pp. 429--436.

\bibitem{20li2018deep}
J.~Li, J.~Feng, and C.-C.~J. Kuo, ``Deep convolutional neural network for
  latent fingerprint enhancement,'' \emph{Signal Processing: Image
  Communication}, vol.~60, pp. 52--63, 2018.

\bibitem{horapong2020progressive}
K.~Horapong, K.~Srisutheenon, and V.~Areekul, ``Progressive latent fingerprint
  enhancement using two-stage spectrum boosting with matched filter and sparse
  autoencoder,'' in \emph{International Conference on Electrical
  Engineering/Electronics, Computer, Telecommunications and Information
  Technology (ECTI-CON)}.\hskip 1em plus 0.5em minus 0.4em\relax IEEE, 2020,
  pp. 531--534.

\bibitem{liu2020automatic}
M.~Liu and P.~Qian, ``Automatic segmentation and enhancement of latent
  fingerprints using deep nested unets,'' \emph{IEEE Transactions on
  Information Forensics and Security}, vol.~16, pp. 1709--1719, 2020.

\bibitem{21dabouei2018id}
A.~Dabouei, S.~Soleymani, H.~Kazemi, S.~M. Iranmanesh, J.~Dawson, and N.~M.
  Nasrabadi, ``{ID} preserving generative adversarial network for partial
  latent fingerprint reconstruction,'' \emph{arXiv preprint arXiv:1808.00035},
  2018.

\bibitem{joshi2019latent}
I.~Joshi, A.~Anand, M.~Vatsa, R.~Singh, S.~D. Roy, and P.~Kalra, ``Latent
  fingerprint enhancement using generative adversarial networks,'' in
  \emph{2019 IEEE Winter Conference on Applications of Computer Vision
  (WACV)}.\hskip 1em plus 0.5em minus 0.4em\relax IEEE, 2019, pp. 895--903.

\bibitem{huang2020latent}
X.~Huang, P.~Qian, and M.~Liu, ``Latent fingerprint image enhancement based on
  progressive generative adversarial network,'' in \emph{IEEE/CVF Conference on
  Computer Vision and Pattern Recognition (CVPR)}.\hskip 1em plus 0.5em minus
  0.4em\relax IEEE, 2020, pp. 8000--801.

\bibitem{23reed2016generative}
S.~Reed, Z.~Akata, X.~Yan, L.~Logeswaran, B.~Schiele, and H.~Lee, ``Generative
  adversarial text to image synthesis,'' \emph{arXiv preprint
  arXiv:1605.05396}, 2016.

\bibitem{24isola2017image}
P.~Isola, J.-Y. Zhu, T.~Zhou, and A.~A. Efros, ``Image-to-image translation
  with conditional adversarial networks,'' in \emph{IEEE conference on CVPR},
  2017, pp. 1125--1134.

\bibitem{RN1275}
I.~Goodfellow, J.~Pouget-Abadie, M.~Mirza, B.~Xu, D.~Warde-Farley, S.~Ozair,
  A.~Courville, and Y.~Bengio, ``Generative adversarial nets,'' in
  \emph{Advances in Neural Information Processing Systems}, vol.~27, pp.
  2672--2680.

\bibitem{2401yeh2017semantic}
R.~A. Yeh, C.~Chen, T.~Yian~Lim, and A.~G. e.~a. Schwing, ``Semantic image
  inpainting with deep generative models,'' in \emph{IEEE Conference on CVPR},
  2017, pp. 5485--5493.

\bibitem{isola2017image}
P.~Isola, J.-Y. Zhu, T.~Zhou, and A.~A. Efros, ``Image-to-image translation
  with conditional adversarial networks,'' in \emph{Proceedings of the IEEE
  conference on computer vision and pattern recognition}, 2017, pp. 1125--1134.

\bibitem{RN761}
A.~Buades, T.~M. Le, J.~M. Morel, and L.~A. Vese, ``Fast cartoon + texture
  image filters,'' \emph{IEEE Transactions on Image Processing}, vol.~19,
  no.~8, pp. 1978--1986, 2010.

\bibitem{25bengio2009learning}
Y.~Bengio \emph{et~al.}, ``Learning deep architectures for {AI},''
  \emph{Foundations and trends{\textregistered} in Machine Learning}, vol.~2,
  no.~1, pp. 1--127, 2009.

\bibitem{27hong2015multimodal}
C.~Hong, J.~Yu, J.~Wan, D.~Tao, and M.~Wang, ``Multimodal deep autoencoder for
  human pose recovery,'' \emph{IEEE Trans. on Image Processing}, vol.~24,
  no.~12, pp. 5659--5670, 2015.

\bibitem{28yu2018multitask}
J.~Yu, C.~Hong, Y.~Rui, and D.~Tao, ``Multitask autoencoder model for
  recovering human poses,'' \emph{IEEE Trans. on Industrial Electronics},
  vol.~65, no.~6, pp. 5060--5068, 2018.

\bibitem{29vincent2008extracting}
P.~Vincent, H.~Larochelle, Y.~Bengio, and P.-A. Manzagol, ``Extracting and
  composing robust features with denoising autoencoders,'' in
  \emph{International Conference on Machine Learning}, 2008, pp. 1096--1103.

\bibitem{30vincent2010stacked}
P.~Vincent, H.~Larochelle, I.~Lajoie, Y.~Bengio, and P.-A. Manzagol, ``Stacked
  denoising autoencoders: Learning useful representations in a deep network
  with a local denoising criterion,'' \emph{Journal of machine learning
  research}, vol.~11, no. Dec, pp. 3371--3408, 2010.

\bibitem{31lu2013speech}
X.~Lu, Y.~Tsao, S.~Matsuda, and C.~Hori, ``Speech enhancement based on deep
  denoising autoencoder.'' in \emph{Interspeech}, 2013, pp. 436--440.

\bibitem{32pathak2016context}
D.~Pathak, P.~Krahenbuhl, J.~Donahue, T.~Darrell, and A.~A. Efros, ``Context
  encoders: Feature learning by inpainting,'' in \emph{IEEE conference on
  CVPR}, 2016, pp. 2536--2544.

\bibitem{39wang2007fingerprint}
Y.~Wang, J.~Hu, and D.~Phillips, ``A fingerprint orientation model based on
  2{D} fourier expansion ({FOMFE}) and its application to singular-point
  detection and fingerprint indexing,'' \emph{IEEE Trans. on PAMI}, vol.~29,
  no.~4, pp. 573--585, 2007.

\bibitem{16mao2016image}
X.-J. Mao, C.~Shen, and Y.-B. Yang, ``Image restoration using convolutional
  auto-encoders with symmetric skip connections,'' \emph{arXiv preprint
  arXiv:1606.08921}, 2016.

\bibitem{44DahlImproving}
G.~E. Dahl, T.~N. Sainath, and G.~E. Hintion, ``Improving deep neural networks
  for {LVCSR} using rectified linear units and dropout,'' 2013, pp. 8609--8613.

\bibitem{37simonyan2014very}
K.~Simonyan and A.~Zisserman, ``Very deep convolutional networks for
  large-scale image recognition,'' \emph{arXiv preprint arXiv:1409.1556}, 2014.

\bibitem{33NISTSD14}
\BIBentryALTinterwordspacing
NIST and FBI. (2010) {NIST} special database {SD}14. [Online]. Available:
  \url{https://www.nist.gov/srd/nist-special-database-14}
\BIBentrySTDinterwordspacing

\bibitem{34MINDTCT}
\BIBentryALTinterwordspacing
NIST. Nbis (nist biometric image software). [Online]. Available:
  \url{https://www.nist.gov/itl/iad/ig/nbis.cfm}
\BIBentrySTDinterwordspacing

\bibitem{49RN403}
R.~Cappelli, D.~Maio, and D.~Maltoni, ``Modelling plastic distortion in
  fingerprint images,'' in \emph{International Conference on Advances in
  Pattern Recognition}, 2001, pp. 371--378.

\bibitem{41NIST}
\BIBentryALTinterwordspacing
NIST and FBI. (2007) {NIST} special database {SD}27. [Online]. Available:
  \url{http://www.nist.gov/itl/iad/ig/sd27a.cfm}
\BIBentrySTDinterwordspacing

\bibitem{43sankaran2015multisensor}
A.~Sankaran, M.~Vatsa, and R.~Singh, ``Multisensor optical and latent
  fingerprint database,'' \emph{IEEE Access}, vol.~3, pp. 653--665, 2015.

\bibitem{42kingma2014adam}
D.~P. Kingma and J.~Ba, ``Adam: A method for stochastic optimization,''
  \emph{arXiv preprint arXiv:1412.6980}, 2014.

\bibitem{RN930}
K.~Cao, D.~L. Nguyen, C.~Tymoszek, and A.~K. Jain, ``End-to-end latent
  fingerprint search,'' \emph{IEEE Transactions on Information Forensics and
  Security}, vol.~15, pp. 880--894, 2020.

\end{thebibliography}

\end{document}